\documentclass[10pt,twocolumn,letterpaper]{article}

\usepackage{cvpr}
\usepackage{times}
\usepackage{epsfig}
\usepackage{graphicx}
\usepackage{amsmath}
\usepackage{amssymb}
\usepackage{flushend}

\usepackage[breaklinks=true,bookmarks=false]{hyperref}

\cvprfinalcopy 

\def\cvprPaperID{****} 

\begin{document}


\title{Hand Segmentation and Fingertip Tracking from Depth Camera Images Using Deep Convolutional Neural Network and Multi-task SegNet}


\author{Duong Hai Nguyen$^\dagger$
	\qquad
	Tai Nhu Do$^\dagger$
	\qquad
	In-Seop Na$^\ddagger$
	\qquad	
	Soo-Hyung Kim$^\dagger$\\\\
	$^\dagger$Chonnam National University\\
	Gwangju, South Korea\\
	$^\ddagger$Chosun University\\
	Gwangju, South Korea\\
	{\tt\small nhduong\_3010@live.com, donhutai@gmail.com, ypencil@hanmail.net, shkim@chonnam.ac.kr}
}

\maketitle

%
%

\begin{abstract}
Hand segmentation and fingertip detection play an indispensable role in hand gesture-based human-machine interaction systems. In this study, we propose a method to discriminate hand components and to locate fingertips in RGB-D images. The system consists of three main steps: hand detection using RGB images providing regions which are considered as promising areas for further processing, hand segmentation, and fingertip detection using depth image and our modified SegNet, a single lightweight architecture that can process two independent tasks at the same time. The experimental results show that our system is a promising method for hand segmentation and fingertip detection which achieves a comparable performance while model complexity is suitable for real-time applications.
\end{abstract}




\section{Introduction}
\label{sec:introduction}
We tend to simplify the way we interact with machines. Instead of controlling a computer mouse, typing on a keyboard, touching a digital pad, or even writing with a pen, current studies are investigating methods which are based on hand gestures. Generally, these systems require a sequence of hand actions or gestures before making decisions according to the input signals. There are two principal ways to acquire user commands: camera-based or equipment-based methods. Regarding the former, thanks to the availability of hi-tech cameras such as Kinect, we can measure not only the discrimination between 2D objects but also depth information which can solve the high-accuracy hand modeling problem. In addition, without wearable devices, camera-based methods are considered as auspicious techniques for use in natural human-machine interaction systems. However, hand segmentation and fingertip detection are still challenging due to the variations in background of the sampling environment. Usually, in order to overcome this issue, researchers work under assumptions about the environmental setup as well as restrictions on users.

To address the above problems, we propose an end-to-end system for hand segmentation and fingertip detection using deep neural networks without any assumptions regarding the sampling environment and user. The purpose of hand segmentation is to discriminate between hand components such as the palm, thumb, index finger, etc. The shapes and locations of these parts can be used to represent hand gestures for further processing. In some cases, fingertip tracing is vital to interacting with machines. Another goal of our work is to develop a practical approach for gesture recognition applications, for example, in the case of BigScreen for Smart Meeting, since not only is the fingertip detection accuracy crucial but also the running-time is of paramount importance.

The rest of the paper is organized as follows: in the second section, we provide an overview on the recent studies related to the stated issue. We then explain our method and the related materials that we used in this research. After that we describe the data and our experimental results. Finally, we conclude our study and discuss future works for further improvements.

\section{Related Works}
\label{sec:related_works}
Regarding the hand component segmentation problem, researchers have focused on two main approaches: wearable device- and image processing-based methods \cite{b1,b2,b3,b4}. While the former does not meet the natural and comfortable interaction criteria, the latter is considered a promising approach with the support of a depth camera \cite{b5,b6,b7}. The authors proposed fitting systems to represent 3D points acquired by Kinect using a hand model. Their methods use both appearance and temporal information to track hand components over time. However, Sharp et al.{\cite{b5}} suffered from multiple-hand tracking issue since they attempted to follow a single object only which is unsuitable for practical applications. Even though Tan et al. {\cite{b6}} showed significant improvements over Sharp et al. {\cite{b5}} by using a detailed mesh personalized to each user, a calibration step is required for a new user to transform a poorly-fit template model into personalized tracking one.

To address the fingertip detection issue, researchers have clearly focused on two main approaches in terms of input signals: the first only uses RGB images while the other considers depth images as well. The latter consistently acquires a reputation for its performance compared to the former since it uses both RGB values and depth signals from the camera. Regarding the first approach, researchers have proposed hand image processing methods based on background subtraction and skin color detection \cite{b8,b9}. However, this approach can be affected by the illumination and variations in the background of the sampling environment, that is, the authors assumed that the images were captured from stationary cameras in a steady scenario. The benefit of this hypothesis is that such simple computer vision techniques can be employed as background subtraction, etc. The other approach for hand image processing, which is to use RGB-D images, was applied in \cite{b10,b11}. In lieu of using a hand detector for RGB images, the authors located fingertips by assuming that the hand must be at the shortest distance from the camera with the given depth information.

In the past few years, deep convolutional neural networks have achieved state-of-the-art performance in numerous computer vision issues. Due to the availability of powerful hardware and public datasets, training deep neural networks is not as difficult or restricting as it was previously \cite{b12}. In this paper, we propose a method for hand segmentation and fingertip detection using RGB-D image and deep neural networks. Our system works well in various sampling scenarios including dynamic and stationary environments, meanwhile, the processing time is up to 15 fps with GPU support. In addition, a calibration step is not indispensable for each user and our method also addressed multiple-hand processing problem. Moreover, another contribution of this paper is a modified version of SegNet for semantic segmentation {\cite{b14}}. Instead of using two different SegNets for multiple tasks including hand component segmentation and fingertip detection, our multi-task SegNet, a single lightweight architecture whose the number of parameters was reduced by $10,014,563$, can process two different tasks at the same time. Last but not least, the modified model performance is similar to the original architecture.

\section{Proposed Method}
\label{sec:proposed_method}
\begin{figure*}[!ht]
	\centering
	\includegraphics[scale=0.43]{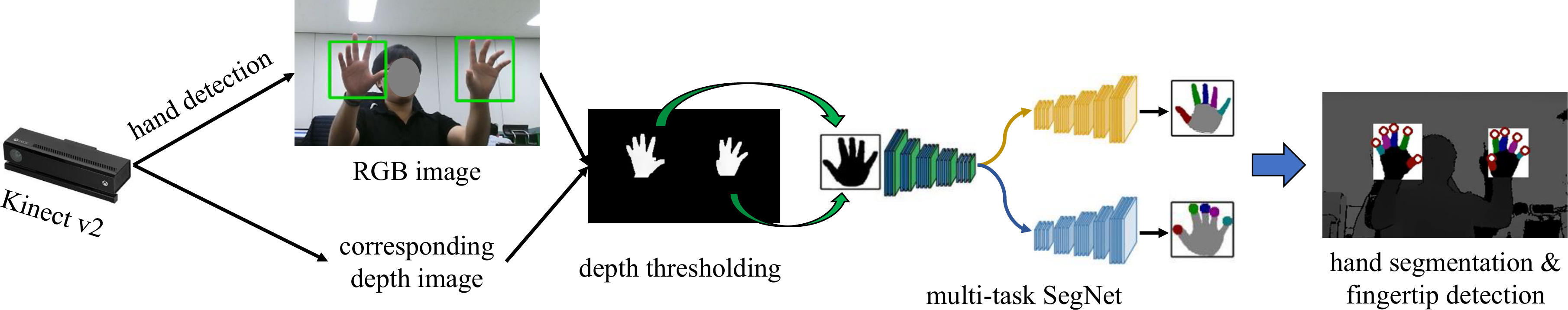}
	\caption{Proposed hand segmentation and fingertip detection pipeline.}
	\label{pro_met}
\end{figure*}

The overview of the proposed method is presented in Fig. \ref{pro_met}.

The system receives RGB and depth images acquired by an RGB-D camera sensor such as Microsoft Kinect V2. We trained YOLOv2 \cite{b13}, a state-of-the-art object detector, from scratch to find the bounding boxes containing hands in RGB signals. The corresponding hand areas in depth image then will be extracted using depth thresholding. For each region of interest in the depth image, we discriminate hand components and identify fingertips using our modified deep autoencoders named multi-task SegNet originally proposed for semantic segmentation {\cite{b14}}. This single model produces two outcomes with respect to hand components and fingertips given a depth image. Note that our system does not require any specific functions from the camera SDK and adapts to several sorts of depth images.

\subsection{Hand Detection}
Hand component segmentation and fingertip detection without providing hand location is a challenging problem to solve. In general, hand components and fingertip regions are tiny compared to others in an image. As a result, finding distinguishable features for them is unachievable in practice. To address this issue, we use YOLOv2 \cite{b13}, which is the current state-of-the-art deep neural network for object detection. Since the conventional YOLOv2 was trained using a dataset which does not take hand images into account, we trained another YOLOv2 from scratch to detect all hand regions in RGB images. The obtained bounding boxes will be used in depth thresholding to extract information in the respective depth image.
\subsection{Depth Thresholding}
\begin{figure*}[!ht]
	\centering
	\includegraphics[scale=0.55]{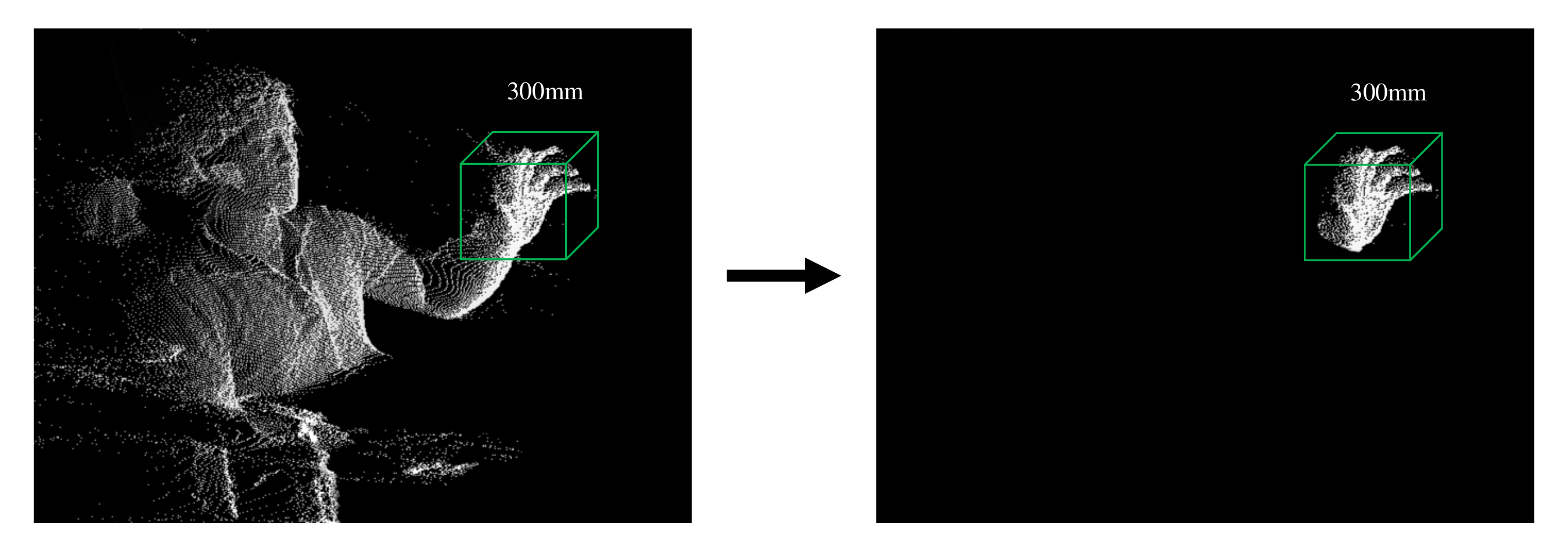}
	\caption{Hand depth thresholding.}
	\label{depth_the}
\end{figure*}

Given two points of the detected hand region ${((x,y)_I,(x',y')_I)}$ from RGB image $I(x,y)$, we extract the hand mask through depth thresholding (Fig. \ref{pro_met}). Assuming that the corresponding depth image is $D(x,y)$, the hand region in the depth image is represented by ${((x,y)_D,(x',y')_D)}$. If $X=(x,x')_D$ and $Y=(y,y')_D$, then $D(X,Y)$ contains depth information of the detected hand region. It can be seen that $m=\text{mode}(D(X,Y))$ is the depth level to recognize hand from the background, therefore, we eliminate all the background pixels using the following formula:
\begin{equation}
	D(x,y)=
	\begin{cases}
		D(x,y),& \text{if } x \in X, y \in Y,\\& \text{and } \|D(x,y)-m\|<t\\
		0, & \text{otherwise}
	\end{cases}.
	\label{depth_thres}
\end{equation}
where $t=300$ is the experimentally-determined threshold for hand binarization. In other words, only pixels inside the rectangular parallelepiped as shown in Fig. \ref{depth_the} are considered as belonging to the hands.

The depth hand image $D(X,Y)$ after removing trivial information will be used as the input for hand segmentation and fingertip detection in our system.

\subsection{Hand Component Segmentation \& Fingertip Detection}
\begin{figure*}[!ht]
	\centering
	\includegraphics[scale=0.33]{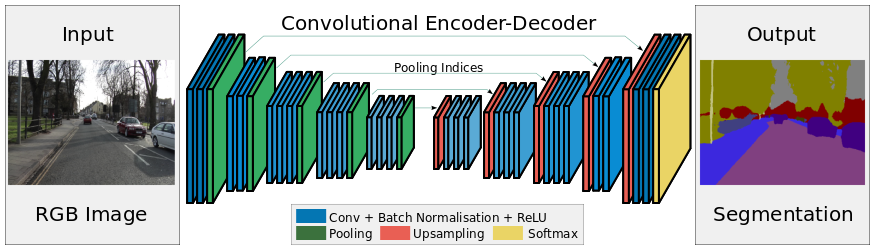}
	\caption{Conventional SegNet for semantic segmentation \cite{b14}.}
	\label{segnet}
\end{figure*}
\begin{figure*}[!ht]
	\centering
	\includegraphics[scale=0.5]{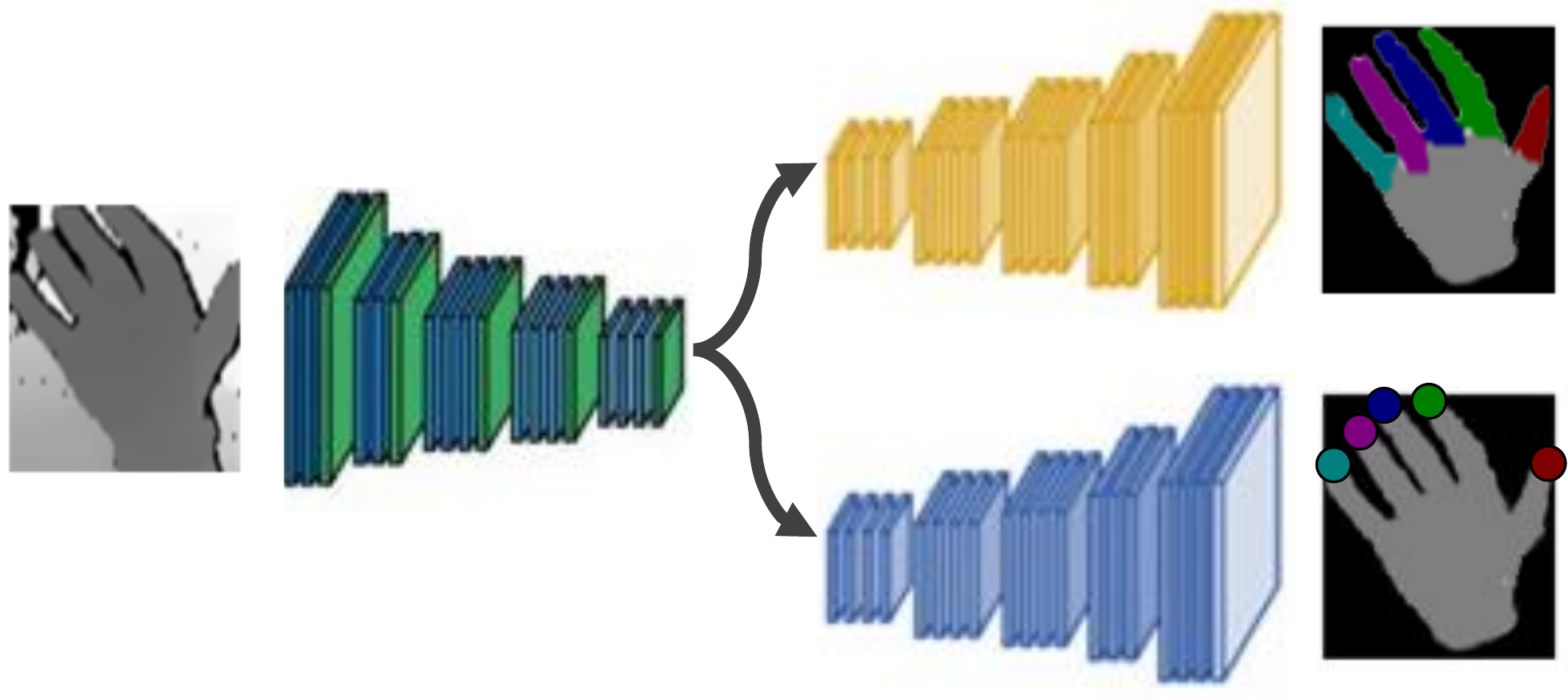}
	\caption{Our multi-task SegNet for hand component segmentation and fingertip detection.}
	\label{7segnet}
\end{figure*}

Given a hand depth image, to solve hand component segmentation and fingertip detection problems, we propose a modified version of SegNet originally used for semantic segmentation {\cite{b14}}. Technically, since the stated issues can be addressed independently, using two different SegNets for multiple tasks is a straightforward option. However, it would make the system heavily depend on high-end GPUs. In general, a conventional SegNet includes an encoder, a corresponding decoder, and a pixel classifier (Fig. {\ref{segnet}}). By feeding an image, conventional SegNet produces an output which includes defined image regions according to training data, e.g., sky, road, vehicle, and so on. In this work, we propose an architecture that combines two identical SegNets to solve multiple tasks at the same time given a depth image as shown in Fig. {\ref{7segnet}}. The backbone of our multi-task SegNet is an encoder with 13 convolutional layers inspired by VGG {\cite{b20}}. Instead of attaching a corresponding decoder to the model, the encoder is followed by two identical decoders. These two network branches are used for hand component segmentation and fingertip detection, respectively. In other words, our proposed model is a single lightweight architecture for solving multiple independent semantic segmentation problems at the same time. As a result, the architecture complexity was reduced by $10,014,563$ parameters while the performance is similar to the conventional version.

By feeding a $96\times96$ hand depth image, the first branch of the network produces seven binary images using feature maps from the encoder which indicate background, palm, and five fingers, respectively. In the second part, using the same feature maps in the encoder, another seven binary images representing background, hand, and five fingertips are generated. The key component of the proposed architecture is sharing. Although the target includes two independent tasks, the same backbone network can be used to achieve a comparable performance while model complexity is suitable for real-time applications.

\section{Experimental Results}
\label{sec:experimentalresults}
\subsection{Datasets}
\begin{figure}[!ht]
	\centering
	\includegraphics[scale=0.58]{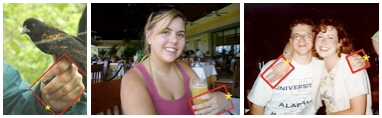}
	\caption{Hand Dataset samples.}
	\label{oxford_hand}
\end{figure}

Originally, YOLOv2 was used for general object detection including human, dog, car, etc. which does not take hands into account. To train the detector from scratch, we use Hand Dataset \cite{b15} which consists of 13,070 labeled RGB hand images for training, validating, and testing. Fig. \ref{oxford_hand} shows sample images of the dataset.

\begin{figure}[!ht]
	\centering
	\includegraphics[scale=0.52]{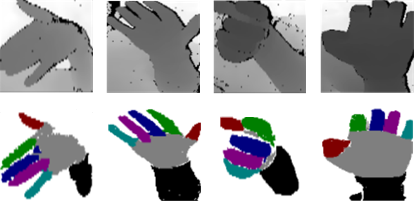}
	\caption{FingerPaint dataset samples with hand component labels: the first row contains depth image while the second one shows the corresponding ground truth.}
	\label{fingerpaint}
\end{figure}

In addition, we train the multi-task SegNet by using both FingerPaint {\cite{b5}} and HandNet {\cite{b16}} datasets. The first dataset was created by five users using Microsoft Kinect camera (Fig. \ref{fingerpaint}) as each of them performed three types of hand poses while capturing. In total, the dataset is divided into 15 parts. We used 70 percent of the dataset for training our model and the rest for testing. Both training and testing sets contain data acquired from five users and three different types of hand poses. Since the original FingerPaint database was not separated, we divided it into different subsets with a ratio of 7:3 for training and testing purposes, respectively.

\begin{figure}[!ht]
	\centering
	\includegraphics[scale=0.55]{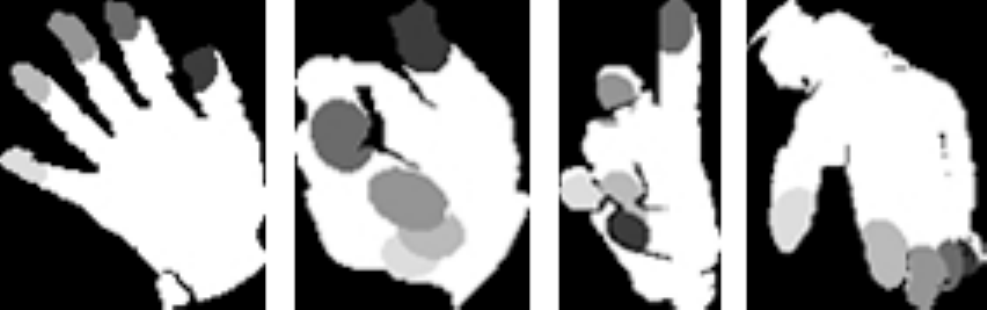}
	\caption{HandNet with fingertip annotations.}
	\label{handnet}
\end{figure}

The HandNet dataset contains a large number of depth images with annotations acquired from RealSense RGB-D cameras of 10 participants. The dataset was divided into three different parts: training, validation, and testing. In this research, we focused on the depth values and the locations of the five fingertips (Fig. \ref{handnet}). It is worth clarifying that our method can adapt to several sorts of depth images acquired by different camera sensors.

\subsection{Experimental Results}
\subsubsection{Implementation}
The multi-task SegNet was trained using Keras {\cite{b17}} with TensorFlow {\cite{b18}} backend for up to 120 iterations in 4 days using one GeForce GTX 1080. The weights were randomly initialized before they were updated by Adam algorithm {\cite{b21}} with mini batch size of 8. The reason for choosing a small batch size in training is due to the limitation in hardware specification. The learning rate started at $10^{-3}$ and the minimum value was $10^{-5}$. To prevent overfitting, data augmentation was performed on the training data. Data preprocessing techniques comprised normalization, translation, rotation, scaling, and mirroring. Regarding the training procedure for YOLOv2, we used default settings that mentioned in {\cite{b13}}. The overall running-time on a GeForce GTX 1080 was around 15 fps which is considered an appropriate speed for real-time applications.

\subsubsection{Hand Component Segmentation}
\begin{figure}[!ht]
	\centering
	\includegraphics[scale=0.45]{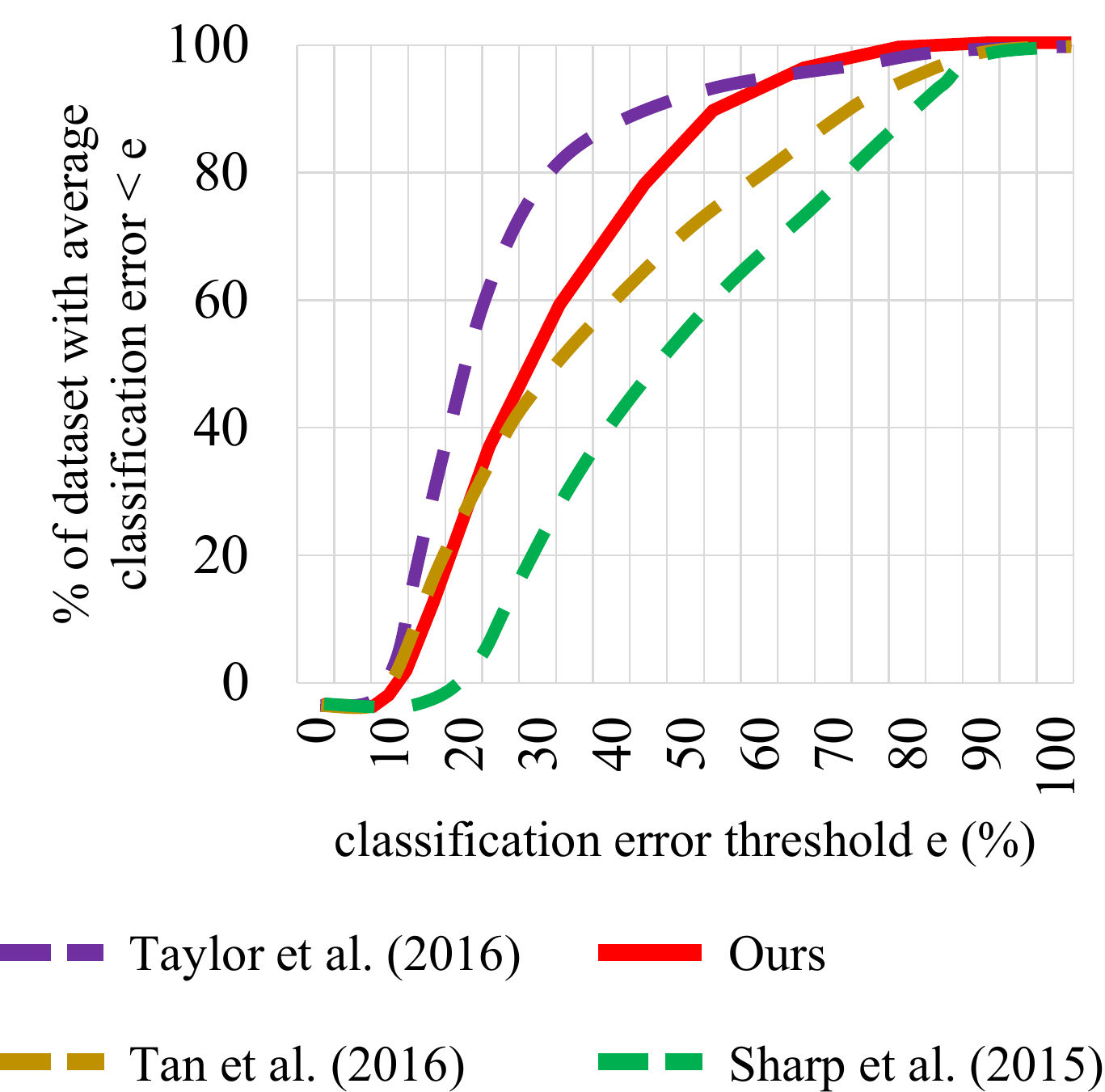}
	\caption{Performance comparison on FingerPaint dataset.}
	\label{fingerpaint_results}
\end{figure}

We compare our method to three other works \cite{b5,b6,b7} in terms of hand component segmentation, as shown in Fig. \ref{fingerpaint_results}.
Our method outperforms the methods described by Sharp et al. \cite{b5} and Tan et al. \cite{b6}. It is also noticed that our method is better than Taylor et al.’s \cite{b7} when the classification error threshold is greater than 60\%. In the range from 10\% to 50\% of the error threshold, it can be seen that our performance is lower than Taylor et al.'s method {\cite{b7}}. We argue the reason is because we did not use personalized items in the FingerPaint dataset {\cite{b5}} for training our model. The running time for hand segmentation is 30fps using GeForce GTX 1080. Moreover, our method can deal with multiple-hand component segmentation without requiring calibration step as mentioned in {\cite{b5,b6}}.

\subsubsection{Fingertip Detection}
\begin{figure}[!ht]
	\centering
	\includegraphics[scale=0.57]{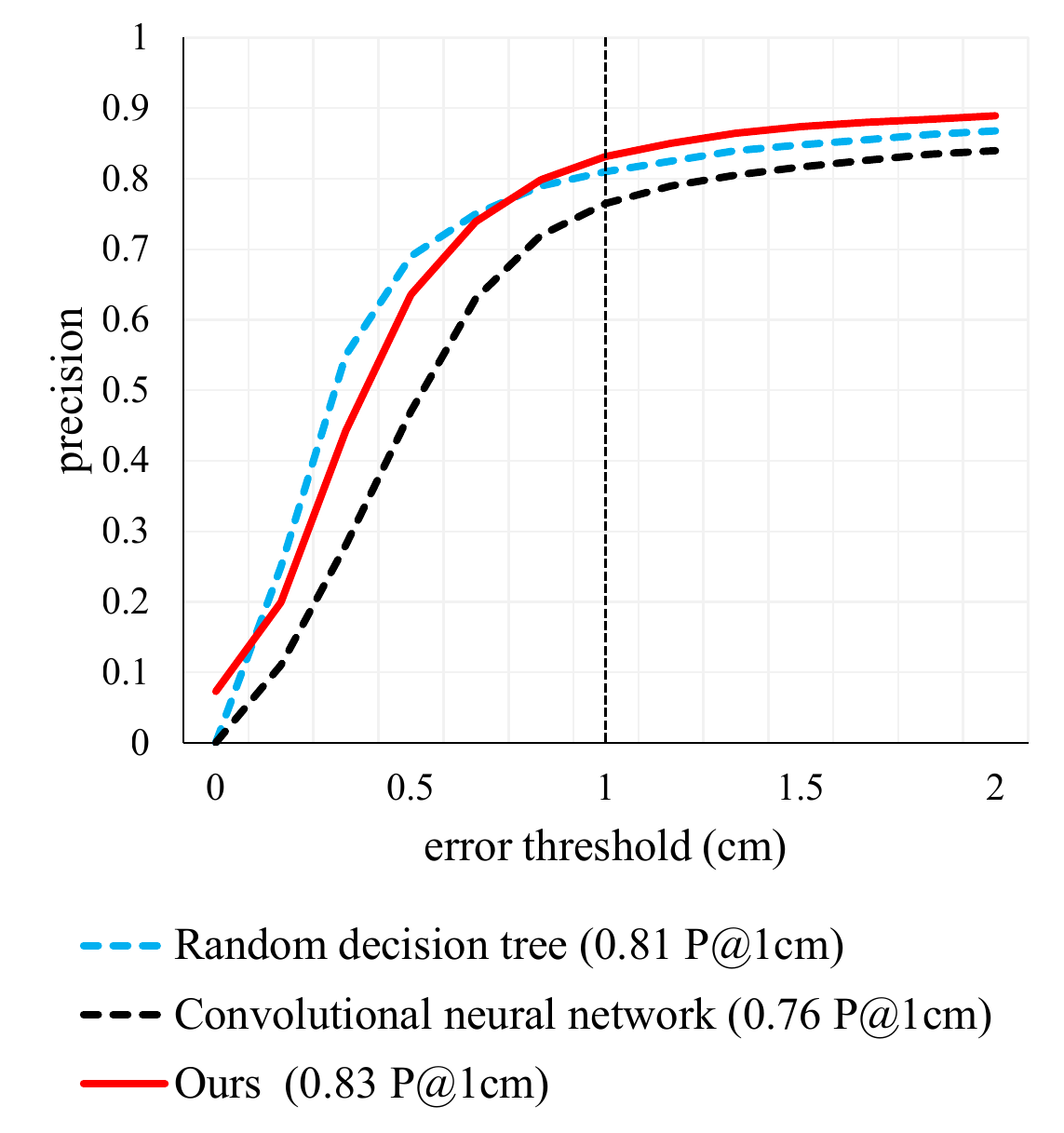}
	\caption{Performance comparison on HandNet dataset.}
	\label{handnet_results}
\end{figure}

In order to evaluate the fingertip detection step, we calculate average detection accuracy with respect to an error threshold and compare to random decision tree and CNN based method {\cite{b16}} using HandNet dataset as shown in Fig. {\ref{handnet_results}}.
Let $e_i$ be the distance (cm) between the predicted location $\hat{p_i}$ and the true fingertip center $p_i$
\begin{equation}
	e_i=\|\hat{p_i}-p_i\|, i=\overline{1,N}.
	\label{error}
\end{equation}
where $N$ is the total number of testing images. By comparing $e_i$ with a small threshold $t_p$, we define the detection precision P as the summation of detection success within a threshold $s_i'$.
\begin{equation}
	P=\sum_{i}s_i'.
	\label{pres}
\end{equation}
where
$s_i'=
\begin{cases}
1, & \text{if } e_i<t_p\\
0, & \text{otherwise}
\end{cases}
.$

Even though our performance is lower than random decision tree in the range from 0.1 to 0.6 of error threshold, for the most part, we use precision at 1.0 error threshold as a benchmark to compare fingertip detection systems on HandNet dataset. Fig. {\ref{handnet_results}} shows that the precision at 1.0 error threshold for the thumb in our system is around 0.83, outperforming two other methods which obtained 0.76 and 0.81.

\begin{figure*}[!ht]
	\centering
	\includegraphics[scale=0.58]{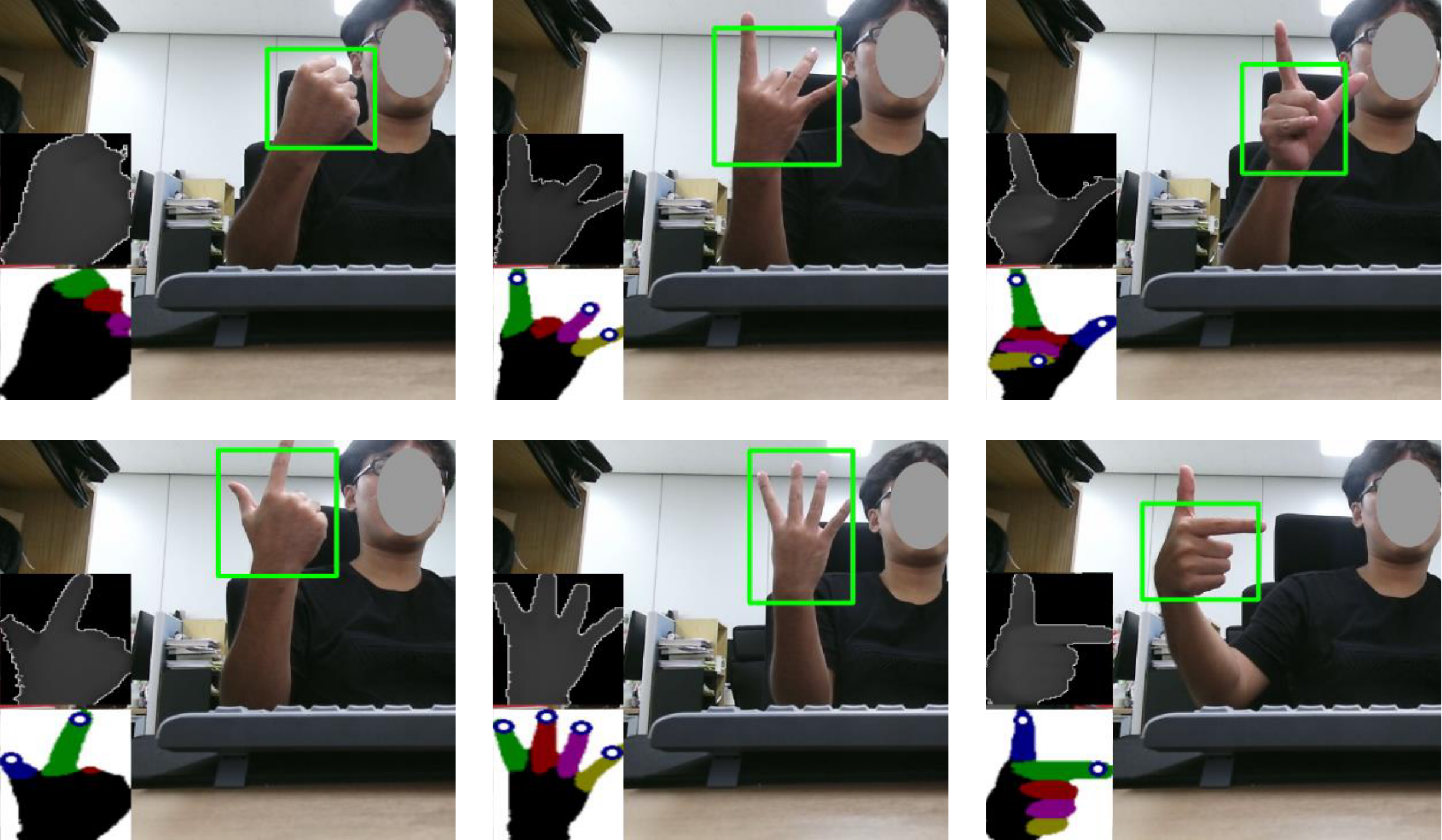}
	\caption{\textit{Local occlusion} in hand component segmentation and fingertip detection.}
	\label{vis}
\end{figure*}

The results support our hypothesis that our modified SegNet not only saves a great number of parameters (up to $10,014,563$) but also achieves a competitive accuracy on FingerPaint and HandNet datasets. The backbone which is a shared encoder might learn a general representation of hands through depth signal while each decoder itself is able to deal with a particular task (i.e., hand component segmentation and fingertip detection) by training from scratch procedure. In addition, our method also addresses \textit{local occlusion problem} (Fig. {\ref{vis}}) as a hand component might be hidden by others.

\section{Conclusion}
\label{sec:conclusion}
In this paper, we introduced a system for hand segmentation and fingertip detection using RGB-D images and deep convolutional encoder-decoder networks. An object detector is used to find hand bounding boxes in RGB signal. Then the corresponding hand areas in depth image are extracted by depth thresholding. In addition, we proposed a multi-task SegNet which is a single lightweight architecture to solve two independent semantic segmentation problems including hand component segmentation and fingertip detection at the same time. The proposed method not only saved a great number of network parameters but also achieved a comparable performance on FingerPaint and HandNet datasets. The experimental results indicate that our method strikes a balance between accuracy and speed for hand gesture recognition-based human-machine interaction systems, which plays a vital role in providing precise and instant feedback to users. For further improvements in terms of the processing time of this system, parallel programming should be taken into account.

\section*{Acknowledgements}
This is supported by Institute for Information \& communications Technology Promotion (IITP) grant funded by the Korea government (MSIT) (No.2017-0-00383, Smart Meeting: Development of Intelligent Meeting Solution based on Big Screen Device). This research was supported by Basic Science Research Program through the National Research Foundation of Korea (NRF) funded by the Ministry of Education (NRF-2017R1A4A1015559).


\bibliographystyle{ieee}
\bibliography{mybibfile}

\end{document}